# Defining Explanation in Probabilistic Systems

Urszula Chajewska
Stanford University
Department of Computer Science
Stanford, CA 94305-9010
*urszula@cs.stanford.edu*

Joseph Y. Halpern
Cornell University
Computer Science Department
Ithaca, NY 14853
*halpern@cs.cornell.edu*

## Abstract

As probabilistic systems gain popularity and are coming into wider use, the need for a mechanism that explains the system's findings and recommendations becomes more critical. The system will also need a mechanism for ordering competing explanations. We examine two representative approaches to explanation in the literature—one due to Gärdenfors and one due to Pearl—and show that both suffer from significant problems. We propose an approach to defining a notion of "better explanation" that combines some of the features of both together with more recent work by Pearl and others on causality.

## 1 INTRODUCTION

Probabilistic inference is often hard for humans to understand. Even a simple inference in a small domain may seem counterintuitive and surprising; the situation only gets worse for large and complex domains. Thus, a system doing probabilistic inference must be able to explain its findings and recommendations to evoke confidence on the part of the user. Indeed, in experiments with medical diagnosis systems, medical students not only trusted the system more when presented with an explanation of the diagnosis, but also were more confident about disagreeing with it when the explanations did not account adequately for all of the aspects of the case (Suermondt and Cooper 1992). Explanation can also play an important role in refining and debugging probabilistic systems. An incorrect or partially correct explanation should be the best indication to an expert of a potential problem.

Our goal is to find a notion of explanation in a probabilistic setting that can be usefully applied by a reasoning system to explain its findings to a human. Of course, we are not the first to examine explanation. It has been has analyzed by philosophers for many years. Traditionally, it has been modeled by introducing a deductive relation between the explanation and the fact to be explained (*explanandum*) (Hempel and Oppenheim 1948). While perhaps applicable to scientific enquiry, this approach is not easily applicable in domains with uncertainty. There have been numerous proposals, both probabilistic and qualitative, for defining explanation in such domains. ((Gärdenfors 1988; Hempel 1965; Salmon 1984) describe the work done by the philosophers and give numerous references; the more recent work in AI includes, for example, (Boutilier and Becher 1995; Henrion and Druzdzel 1990; Pearl 1988; Shimony 1991; Suermondt 1992).) Since we are interested in explanation in probabilistic systems, our focus is on proposals that seek a probabilistic connection between the explanation and the explanandum. In the philosophical literature, the focus has been on the probability of the explanandum given the explanation. The requirements range from just requiring that this conditional probability change, to requiring that it be very high, to requiring that it be greater than the unconditional probability of the explanandum (so that learning the explanation increases the probability of the explanandum); see (Gärdenfors 1988; Salmon 1984) for discussion and further references. In contrast, the research on explanation in Bayesian networks (Henrion and Druzdzel 1990; Pearl 1988; Shimony 1991) has concentrated on computing the conditional probability of the explanation given the explanandum, adding in some cases the additional requirement that the explanation be a complete world description.

Clearly the appropriateness of a notion of explanation will depend in large part on the intended application. A scientific explanation might well have different properties from an explanation provided by an intelligent tutoring system. In our intended application, the system will typically have some uncertainty regarding the true state of the world (and possibly even the domain's causal structure), represented as a probability distribution. Note that this is different from, say, an intelligent tutoring system, where we assume the system to have the full knowledge of the domain. For simplicity, we make the (admittedly unrealistic) assumption that the user's knowledge can be identified with the system's knowledge.[1] Because we expect that there will typically be a number of competing explanations that can be provided to the user, we are interested not just in finding an absolute notion of explanation, but a comparative notion. We want to be able to judge when one explanation is better than another.

---

[1] Modeling the user's knowledge and adjusting the explanation to fit it is one of the planned extensions of this work.

In this paper, we concentrate on two definitions of explanation, one due to Gärdenfors (1988) and the other to Pearl (1988), as representatives of the two approaches mentioned above. While, as we point out, there are significant problems with these definitions, we consider them because they have some important features that we feel should constitute part of an approach to defining explanation. We suggest an approach that combines what we feel are the best features of these two definitions with some ideas from the more recent work on causality (Balke and Pearl 1994; Druzdzel and Simon 1993; Heckerman and Shachter 1995; Pearl 1995).

One of the observations that falls naturally out of our approach is that we should expect different answers depending on whether we are asking for an explanation of beliefs or facts. For example, if the agent believes that it rained last night and we ask for an explanation for this belief, then a perfectly reasonable explanation is that he or she noticed the wet grass in the morning, which is correlated with rain. However, if the agent observes that it is raining and we ask for the best explanation of this observation, then it would certainly not be satisfactory to be told that the grass is wet. We do not accept the wet grass as an explanation in the second case because the wet grass is not a cause of rain. However, we would accept it in the first case because the agent *believing* that the grass is wet is a cause of the agent *believing* that it rained. The critical difference between explanations of beliefs and explanations of observations does not seem to have been discussed before in the literature.

The rest of the paper is organized as follows. In Sections 2 and 3 we present and analyze Gärdenfors' and Pearl's definitions. In Section 4 we present a new approach which generalizes elements of both. We conclude with some open problems in Section 5.

## 2 GÄRDENFORS' APPROACH

### 2.1 THE DEFINITION

As we suggested earlier, roughly speaking, for Gärdenfors, $X$ is an explanation of $E$ if $\Pr(E|X) > \Pr(E)$. That is, $X$ is an explanation of $E$ if learning $X$ raises the probability of $E$. In order to flesh out this intuition, we need to make precise what probability distribution we are using.

According to Gärdenfors, what requires explanation is something that is already known, but was unexpected: A person asking for an explanation expresses a "cognitive dissonance" between the explanandum and the rest of his or her beliefs. We don't typically require an explanation for something we expected all along. The amount of dissonance is measured by the surprise value of the explanandum in the belief state in which we reject our belief in the explanandum while holding as many as possible of our other beliefs intact (this operation is called *contraction* and comes from the belief revision framework (Alchourrón, Gärdenfors, and Makinson 1985)). An explanation provides "cognitive relief"; the degree of "cognitive relief" is measured by the degree to which the explanation decreases the surprise value.

For example, if we ask for an explanation of why David has the flu, then we already know that David has the flu. Thus, if $E$ is the statement "David has the flu", then in the current situation, we already ascribe probability 1 to $E$. Nothing that we could learn could increase that probability. On the other hand, we presumably asked for an explanation because before David got sick, we did not expect him to get sick. That is, if $\Pr_E^-$ describes the agent's probability distribution in the contracted belief state, before David got flu, we expect $\Pr_E^-(E)$ not to be too high. An explanation $X$ (like "David was playing with Sara, who also has the flu") would raise the probability of $E$ in the contracted belief state, that is, we have

$$\Pr_E^-(E|X) > \Pr_E^-(E).$$

As Gärdenfors' definition stresses, what counts as an explanation depends on the agent's epistemic state. An explanation for one agent may not be an explanation for another, as the following example, essentially taken from (Gärdenfors 1988), shows.

**Example 2.1** *If we ask why Mr. Johansson has been taken ill with lung cancer, the information that he worked in asbestos manufacturing for many years is not going to be a satisfactory explanation if we don't know anything about the effects of asbestos on people's health. Adding the statement "70% of those who work with asbestos develop lung cancer" makes the explanation complete. The explanation must consist of both statements. However, if we try to explain Mr. Johansson's illness to his close friend, who is likely to know his profession, we would supply only the second piece of information. Similarly, to someone who knows more about asbestos but less about Mr. Johansson, we would only present the information about his profession.*

To formalize these intuitions, Gärdenfors characterizes a (probabilistic) epistemic state using the possible worlds model. At any given time, an agent is assumed to consider a number of worlds (or states of the world) possible. For example, if the agent looks out the window and notices that it is raining, his set of possible worlds would include only worlds where it is raining. Learning new facts about the world further restricts the set of the worlds we consider possible. Among the possible worlds, some may be more likely than others. To describe this likelihood, the agent is assumed to have a probability distribution over the possible worlds.

Thus, an epistemic state is taken to be a pair $K = \langle W, \Pr \rangle$, where $W$ is a set of possible worlds (or possible states of the world) and $\Pr$ is a probability distribution on $W$. A sentence $A$ is said to be *accepted as knowledge* in an epistemic state $K$ if $\Pr(A) = 1$. We sometimes abuse notation and write $A \in K$ if $A$ is accepted in epistemic state $K$.

Given an epistemic state $K = \langle W, \Pr \rangle$ of an agent, let $K_E^- = \langle W_E^-, \Pr_E^- \rangle$ denote the *contraction* of $K$ with respect to $E$, i.e., the epistemic state characterizing the agent's beliefs that is as close to $K$ as possible such that $E \notin K_E^-$. Gärdenfors describes a number of postulates that $K_E^-$ should satisfy, such as $K_E^- = K$ if $E \notin K$. It is beyond the scope of this paper to discuss these postulates (see (Alchourrón, Gärdenfors, and Makinson 1985)). However, these postulates do not serve to specify $K_E^-$ uniquely; that is, given $K$ and $E$, there may be several epistemic states $K'$ that satisfy the postulates. On the other hand, there are some situations where it is straightforward to specify $K_E^-$. For example, if Pr is determined by a Bayesian network together with some observations, including $E$, then $\Pr_E^-$ is just the distribution that results from the Bayesian network and all the observations but $E$.

We can now present Gärdenfors' definition of explanation.

**Definition 2.2** (from (Gärdenfors 1988)) $X$ *is an explanation of $E$ relative to a state of belief $K = \langle W, \Pr \rangle$ (where $E \in K$) if*

1. $\Pr_E^-(E|X) > \Pr_E^-(E)$, *and*

2. $\Pr(X) < 1$ *(that is, $X \notin K$).*

We have already seen the first clause of this definition. Note that, in this clause (and throughout this paper), we identify the formulas $E$ and $X$ with sets of possible worlds, namely, the sets of worlds (in $W_E^-$) in which $E$ and $X$, respectively, are true. The second clause helps enforce the intuition that the explanation depends on the agent's epistemic state. The explanation cannot be something the agent already knows. For example, fire will not be an explanation of smoke if the agent already knows that there is a fire. Notice that the second clause also prevents $E$ from being an explanation for itself. (Clearly $E$ satisfies the first clause, since $\Pr_E^-(E|E) > \Pr_E^-(E)$; since we have assumed $E \in K$, $E$ does not satisfy the second clause.) Unfortunately, as we shall see, while the second clause does exclude $E$ as an explanation, it does not exclude enough.

Given this notion of explanation, we can define an ordering on explanations that takes into account the degree to which an explanation raises the probability of the explanandum. Gärdenfors in fact defined *explanatory power* as the difference between the posterior and prior probability of the explanandum. Thus, a better explanation is one with better explanatory power.

The difference is not always a good measure of distance between probabilities. An explanation which raises the probability of a statement of interest from 0.500001 to 0.51 is not so powerful. On the other hand, an explanation raising the probability from 0.000001 to 0.01 would be received quite differently, although the difference in probabilities is the same. A more natural way to define explanatory power is by using the ratio of the two probabilities.

**Definition 2.3** *The* explanatory power (EP) *of $X$ with respect to $E$ is*

$$EP(X, E) = \frac{\Pr_E^-(E|X)}{\Pr_E^-(E)}.$$

According to this definition, the two explanations above have dramatically different explanatory power. For this paper, we take the latter definition as our formal definition of explanatory power.

Before we get to our critique of Gärdenfors' definition, there is one other issue we need to discuss: the language in which explanations are given. Definition 2.3 makes perfect sense if, for example, explanations are propositional formulas over a finite set of primitive propositions. In that case, a world $w$ could be taken to be a truth assignment to a finite family of these primitive propositions. We could also take explanations to be first-order formulas, in which case a world could be taken to be a first-order interpretation. Gärdenfors in fact allows even richer explanations, involving statistical statements. As we saw, in Example 2.1, a possible explanation of Mr. Johansson's illness for someone who already knew that he worked in asbestos manufacturing is to say "70% of those who work with asbestos develop lung cancer". To make sense of this, Gärdenfors associates with a world not only a first-order interpretation, but a distribution over individuals in the domain. (This type of model is also considered in (Halpern 1990), where a structure consists of possible worlds, with a distribution over the worlds, and, in each world there is a distribution on the individuals in that world; a formal language is provided for reasoning about such models. If the domain is finite, we could simplify things and assume that the distribution is the uniform distribution, as is done in (Bacchus, Grove, Halpern, and Koller 1996).) While it is not necessary to consider such a rich language to make sense of Gärdenfors' definition, one of his key insights is that statistical assertions are an important component of explanations. Indeed, he explicitly describes an explanation as a conjunction $X_1 \wedge X_2$, where $X_1$ is a conjunction of statistical assertions and $X_2$ is what Gärdenfors calls a *singular* sentence, by which he means a Boolean combination of atomic sentences in a first-order language with only unary predicates. (Either conjunct may be omitted.) As we shall argue, we need to generalize this somewhat to allow causal assertions as well as statistical assertions.

### 2.2 A CRITIQUE

While Gärdenfors' definition has some compelling features (see (Gärdenfors 1988) for further discussion), it also has some serious problems, both practical and philosophical. We describe some of them in this section.

1. While the second clause prevents $E$ from being an explanation of itself, there are many other explanations that it does not block. Let $F$ be any formula such that $\Pr(F) < 1$ and $\Pr_E^-(E \wedge F) > 0$. Then $E \wedge F$ will be an explanation for $E$. Moreover, it will be the explanation with the highest possible explanatory power (both according to Gärdenfors' original definition and our modification). This is obvious, since $\Pr_E^-(E|E \wedge F) = 1$. Note that $F$ can be practically any formula here. We surely wouldn't want to accept "$E$ and the coin lands heads" as an explanation for $E$. One possible solution to this problem is to restrict explanations to only involving certain propositions. For example, if we are looking for an explanation for some symptoms, we might require that the explanation be a disease. There are many cases where such restrictions make sense, but if we are to do this, then we must explain where the restrictions are coming from.

2. Even if we restrict attention to a particular vocabulary for explanations, there is nothing preventing us from adding irrelevant conjuncts to an explanation. More precisely, note that if $X$ is an explanation of $E$, and $C$ is conditionally independent of $E$ given $X$, then $\Pr_E^-(E|X) = \Pr_E^-(E|X \wedge C)$. Thus, $X$ and $X \wedge C$ are viewed as equally good explanations.

3. The definition does not take into account the likelihood of the explanation. For example, suppose there are two explanations for a symptom $s$, disease $d_1$ and disease $d_2$, with the same explanatory power, but $d_1$ is a relatively common disease, while $d_2$ is quite rare. If the explanation is given by an expert that is trusted by the user (as in the case of an intelligent tutoring system), then once we are told that, say $d_2$ is the explanation, we would presumably accept it as true. In this case, the prior probability (i.e., the fact that $d_2$ is rare) is irrelevant. However, in our context, even if $\Pr_s^-(s|d_1) = \Pr_s^-(s|d_2)$, it seems clear that we should prefer the explanation $d_1$ to $d_2$.

4. The fact that learning $X$ raises the probability of $E$ does not by itself qualify $X$ to be an explanation of $E$. For example, suppose $s$ is a symptom of disease $d$ and Bob knows this. If Bob learns from a doctor that David has disease $d$ and asks the doctor for an explanation, he certainly would not accept as an explanation that David has symptom $s$, even though $\Pr_d^-(d|s) > \Pr_d^-(d)$. Gärdenfors is aware of this issue, and discusses it in some detail (1988, p. 205). He would call $s$ an explanation of $d$, but not a *causal* explanation. Gärdenfors provides a definition of causal explanation. Unfortunately, while it deals with this problem, it does not deal with the other problems we have raised, so we do not discuss it here.[2] We disagree with Gärdefnors that there are explanations that are not causal; we view all explanations as causal. In particular, we do not think that Bob would accept $s$ as an explanation of $d$ at all. Note, however, that if Bob had asked the doctor why he (the doctor) believed that David had disease $d$, an acceptable explanation would have been that the doctor believed (or knew) that David had symptom $s$. There is a big difference between what Bob would accept as an explanation for $d$ and what he would accept as an explanation of the doctor's belief that $d$. We return to this issue below.

5. As a practical matter, Gärdenfors' definition requires the computation of the contraction of a belief state (besides the computation of many conditional probabilities in that contracted belief state). If an approach like this is to be used in a system, we need techniques for computing the contraction. More accurately, since the contraction is not unique, we need to focus on applications where there is a relatively straightforward notion of contraction.

## 3  THE MAXIMUM A POSTERIORI MODEL APPROACH

### 3.1  THE DEFINITION

Most of the work done on explanation in belief networks was based on the intuition that the best explanation for an observation is the state of the world that is most probable given the evidence (Henrion and Druzdzel 1990; Pearl 1988; Shimony 1991). There is no notion of "cognitive dissonance" or surprise. The explanation is an (informed) guess about the possible world we are currently in, based on the evidence (which includes the explanandum). In some cases (e.g., (Pearl 1988)), the guess must specify the world completely—formulas describing sets of worlds are not allowed as explanations. This approach, which we call *Maximum A Posteriori* model (MAP) after (Shimony 1991), has been also known under other names: *Most Probable Explanation (MPE)* (Pearl 1988) and *Scenario-Based Explanation* (Henrion and Druzdzel 1990).

Formally, according to Pearl, given an epistemic state $K = \langle W, \Pr \rangle$, an explanation for $E$ is simply a world $w$ in which $E$ is true. This notion of explanation induces an obvious ordering on explanations. World $w_1$ is a better explanation of $E$ than $w_2$ if $E$ is true in both $w_1$ and $w_2$ and $\Pr(w_1|E) > \Pr(w_2|E)$. Finally, the *best* or *most probable* explanation (MPE) is the world $w^*$ such that $\Pr(w^*|E) = \max_{w \in W} \Pr(w|E)$.[3]

We remark that although we have spoken here of an explanation as being a world, we could equally well take an explanation to be the formula that characterizes the world if we assume (as Pearl does) that each world is uniquely characterized by a formula. If our vocabulary consists of a finite number of propositions $p_1, \ldots, p_k$, and each world is a truth assignment to these primitive propositions, then an explanation would have the form $q_1 \wedge \ldots \wedge q_k$, where each $q_i$ is either $p_i$ or $\neg p_i$. Of course, if we have richer languages, finding formulas that characterize worlds becomes more of an issue.

Two other variants of the MAP approach have been proposed, by Henrion and Druzdzel (1990) and Shimony (1991, 1993). They share with Pearl's definition two important features: First, the explanation is a truth assignment to a subset of propositions, *including* the explanandum. Second, the ordering of explanations is based on their posterior probability given the explanandum.

---

[2] For the interested reader, $C$ is said to be a *causal explanation* of $E$ with respect to belief state $K$ such that $E \in K$ if (1) $\Pr(C) < 1$, (2) $\Pr_E^-(E|C) > \Pr_E^-(E)$, (3) $(\Pr_C^+)_C^- = \Pr_E^-$, where $\Pr_C^+$ is the belief state that arises when we add $C$ to the stock of beliefs in $K$. This is the notion called *belief expansion* (Alchourrón, Gärdenfors, and Makinson 1985). Thus, we add clause (3) to the definition of explanation. Note, however, if $F$ is independent of $E$, then $E \wedge F$ would be a causal explanation of $E$. Similar arguments show that Gärdenfors' definition of causal explanation still suffers from all the other problems we have raised.

[3] Actually, Pearl did not define the notion of explanation, just that of most probable explanation. However, our definitions are certainly in the spirit of his. Also, he did not talk explicitly of worlds and epistemic states, but these are implicit in his definitions. Pearl assumes that there is a Bayesian network that describes a number of variables of interest. The set $W$ then consists of all possible assignments to the variables, and the probability distribution Pr on $W$ is determined by the Bayesian network.

Henrion and Druzdzel actually discuss a number of approaches to explanation. Of most relevance here are what they call *scenario-based explanations*. They assume a tree of propositions (a *scenario tree*), where a path from the root to a leaf represents a scenario, or a sequence of events. They are looking for the scenario with the highest probability given the explanandum.[4] Thus, their approach differs from Pearl's in that the system has additional knowledge (the scenarios). They also allow explanations to be *partial*. The truth values of all propositions do not have to be specified. However, explanations are restricted to coming from a set of prespecified scenarios.

Shimony (1991, 1993) also allows partial explanations. He works in the framework of *Bayesian networks* (as does Pearl, in fact, although his definition makes sense even if probabilities are not represented using Bayesian networks).[5] In his framework, the explanandum is an instantiation of (truth assignment to) some nodes in the network; these are called the *evidence nodes*. An explanation is a truth assignment to the "relevant" nodes in the network. The relevant nodes include the evidence nodes and only ancestors of evidence nodes can be relevant. Roughly speaking, an ancestor of a given node is irrelevant if it has the property that it is independent of that node given the values of the other ancestors. In (Shimony 1991), the best explanation is taken to be the one with the highest posterior probability. In (Shimony 1993), this is extended to allow explanations to be sets of partial truth assignments, subject to certain constraints (discussed in more detail in Section 4.3.)

## 3.2   A CRITIQUE

The MAP approach has an advantage over Gärdenfors': it doesn't require contraction. However, it has its own problems. Some of the problems are particularly acute in Pearl's approach, with its requirement that the explanation be *complete*; i.e., a world; they are alleviated somewhat if we allow partial explanations (sets of worlds). However, some of the problems arise in all variants of the approach, and are a consequence of ordering according to the posterior probability distribution.

1. By making the explanation a complete world, the notion becomes very sensitive to the choice of language, as Pearl himself observes.[6] For example, if our language consists of $\{s, d_1, d_2\}$, then the best explanation for symptom $s$ might be $d_1$, or, more precisely, the world characterized by $s \wedge d_1 \wedge \neg d_2$. For the purposes of this example, suppose that diseases are mutually exclusive, so all worlds where the agent has more than one disease have probability 0. Now suppose we subdivide $d_1$ into two diseases $d_1'$ and $d_1''$, again mutually exclusive (as, for example, hepatitis can be subdivided into hepatitis A and hepatitis B). Then we might find $d_2$ to be a better explanation than either $d_1'$ or $d_1''$ (that is, $\Pr(\neg d_1' \wedge \neg d_1'' \wedge d_2|s)$ may be greater than either $\Pr(d_1' \wedge \neg d_1'' \wedge \neg d_2|s)$ or $\Pr(\neg d_1' \wedge d_1'' \wedge \neg d_2|s)$).

   Pearl gives an even sharper example of this phenomenon. Suppose that $d_1$ is a a diagnosis of perfect health, $d_2$ is a diagnosis of a fatal disease, $\Pr(d_1|s) = 0.8$, and $\Pr(d_2|s) = 0.2$. Now suppose we expand the vocabulary to include $h_1, \ldots, h_8$, where the $h_i$'s are possible holidays that the agent will take next year (provided he or she is indeed healthy), and the agent considers each of these vacation plans equally likely. Then we have that $\Pr(h_i \wedge d_1|s) = 0.1$, and the most likely explanation of the symptom has changed from $d_1$ to $d_2$! So just by considering possible holidays he might take given that he is healthy, the agent finds that the best explanation for his symptoms becomes a fatal disease.

2. A related problem is the fact that if we have a large number of primitive propositions, most will probably be irrelevant or only marginally relevant to explaining a particular proposition. Yet, Pearl's definition forces us to consider worlds, thus forcing us to worry about the truth value of all propositions. This can cause computational problems. In addition, conciseness is a desirable feature in an explanation, particularly in an interactive system. The user usually wants to know only the most influential elements of the complete explanation, and does not want to be burdened with unnecessary detail. This problem is particularly severe if we insist on complete explanations. However, Shimony's partial explanations are not necessarily as concise as one would hope either. It is not hard to show that for each evidence node $X$, the explanation must include an assignment to all the nodes in at least one path from $X$ to the root, since for each relevant node, at least one of its parents must be relevant. Moreover, the irrelevance condition is quite strong and only in limited contexts is it likely to achieve significant pruning. Shimony attempts to overcome this problem by relaxing the irrelevance assumption to what he calls approximate or $\delta$-irrelevance. While helpful in some domains, the extent to which it will result in concise explanations in general is not clear. We discuss this point in more detail in the full paper.

3. The ordering on explanations used in the MAP approach is supposed to maximize the probability of the explanation given the explanandum. However, if we consider only explanations which include the explanandum (as all MAP explanations do), this reduces to maximizing the prior of the explanation. The ordering is then based only on the likelihood of the explanation and not in any way on the degree to which the explanation raises the probability of the explanandum.

---

[4] Actually, they suggest presenting all scenarios that have sufficiently high probability, and pointing out how the most probable one differs from the other likely scenarios.

[5] Recall that a Bayesian network is an acyclic directed graph whose nodes represent primitive propositions (or random variables), together with conditional probability tables describing the probability of a node given instantiations of its parents (Pearl 1988).

[6] Shimony (1991) calls this the *overspecification* problem.

4. All the MAP approaches discussed above consider essentially propositional languages. Once we move to richer languages (like first-order, or languages that allow statistical information), then each world may end up having very low probability. Indeed, if we have a continuous number of worlds, each world may have probability 0. In this case, the definition which requires explanations to be complete worlds is not even useful.

Given the difficulties with complete explanations, why do Shimony and Henrion and Druzdzel put such restrictions on the allowable partial explanations? What is the problem with partial explanations? Suppose our language consists of the propositions $\{p_1, \ldots, p_k\}$. Why not just allow $p_1$ as an explanation, instead of requiring something like $p_1 \land \neg p_2 \land \ldots \land \neg p_k$? Gärdenfors certainly allows such explanations. It is not hard to see why Pearl does not allow partial explanations. Notice that a partial explanation is really a set of worlds (or equivalently, the disjunction of the formulas representing the worlds). But a disjunction will always have higher conditional probability than any of its disjuncts (except in the degenerate case where all but one of the disjuncts has probability 0), and thus will be viewed as a more probable explanation than any of its disjuncts. It is because of this that Shimony puts restrictions on the allowable partial explanations as well. As we shall see, we can deal with this problem, at least to some extent, by modifying the ordering of explanations.

## 4 SYNTHESIS

As we have seen, both Gärdenfors' definition and the MAP definition have problems. We believe that in order to deal with these problems, we need to deal with two relatively orthogonal issues: (1) we must decide what counts as an explanation, and (2) we must decide how to compare two explanations.

### 4.1 WHAT COUNTS AS AN EXPLANATION?

The MAP approach seems somewhat too restrictive in what counts as an explanation: An explanation must be a complete description of a world (or a restricted form of partial explanation). Gärdenfors, on the other hand, is not restrictive enough. He allows $E \land C$ to be an explanation of $E$, for example, and this seems to us unreasonable. In addition, he would allow a falling barometer reading to be an explanation for a storm, thus missing out on the causal structure.

As we mentioned above, we view all explanations as causal. We distinguish between explaining facts and explaining beliefs, but in both cases we look for the same thing in an explanation: a causal mechanism which (possibly together with some facts) is responsible for the fact observed or the beliefs adopted. By enforcing causality, AGMwe believe that we can avoid the problems in Gärdenfors' definition, while still allowing more general explanations than the MAP approach would allow. We remark that we are not the first to stress the role of causality in explanation. Salmon (1984) discusses the issue at length, although the technical details of his proposal are quite different from ours.

The literature on causality is at least as large as the literature on explanation; it is well beyond the scope of this paper to develop a new theory of causality. For the purposes of the rest of this paper, we work at the propositional level (since that is essentially what the recent approaches to causality do) and assume that the causal mechanism is described by a *causal structure*, which we take to be a Bayesian network interpreted causally.

We believe that much of what we do is independent of the particular way we choose to model causality. In particular, we can replace the causal network by structural equations, as described in (Druzdzel and Simon 1993; Pearl 1995). We have chosen to use Bayesian networks as our representation for causality simply to make it easier to relate our approach to Pearl's approach.

In this setting, part of the agent's uncertainty concerns what the right causal mechanism is. For example, an agent may be uncertain whether smoking causes cancer or whether there is a gene that causes both a susceptibility to cancer and a susceptibility to smoking. Thus, we assume that a world is a pair $(w, C)$ consisting of a truth assignment $w$ and a causal structure $C$. As before, an epistemic state $K$ is a pair $\langle W, \text{Pr} \rangle$, where $W$ is a set of worlds of this form, and Pr is a probability distribution on $W$. However, we assume that this epistemic state arises from a simpler description: We assume that the agent has a probability distribution Pr$'$ on causal structures and has made some observations. Notice that a causal structure $C$ also places a probability distribution Pr$_C$ on worlds. We require that the distribution Pr be consistent with the causal mechanisms considered possible and the observations $O$ in the following sense: There must be a probability distribution Pr$'$ on causal mechanisms such that $\Pr(w, C) = \Pr'(C)\Pr_C(w|O)$: that is, the probability of $(w, C)$ is the probability of the causal mechanism $C$ times the probability that $C$ induces on $w$, given the observation. In particular, this means that if the agent considers only one causal mechanism possible, we can identify Pr with a probability on truth assignments, just as Pearl does.

We assume that the explanandum $E$ is one of the observations. This means that $\Pr_E^-$ has a simple form: $\Pr_E^-(w, C) = \Pr'(C)\Pr_C(w|O - \{E\})$. It is easy to see that this definition satisfies the postulates for contraction. An *explanation* of $E$ in epistemic state $K$ is a conjunction $X = X_1 \land X_2$ consisting of a *partial causal mechanism* $X_1$ (that is, a description of a causal structure; see below) and an instantiation of nodes $X_2$ that causally precede $E$ in $X_1$ such that $\Pr(X) < 1$. (We defer for now the issue of whether the explanation raises the probability of the explanandum.)

We are deliberately being vague about the language used to describe the causal mechanism, since we believe that this is an area for further research. For the purposes of this paper, we can take $X_1$ to be simply a description of a subgraph of the causal graph (intuitively, that part of the causal graph that is relevant to explaining $E$, i.e., a subset of the set of paths from nodes in $X_2$ to $E$).

We allow the conjunct describing the causal mechanism to be missing from the explanation if it is known. (In practice, this might mean that the system providing the explanation believes that the agent to whom the explanation is being provided knows the causal mechanism.) Notice that if the agent knows the causal mechanism, and thus considers only one causal mechanism possible (as is implicitly the case when a situation is described by a Bayesian network which is given a causal interpretation), then a world can be identified with a truth assignment. In this case (ignoring the requirement that all the conjuncts in a basic explanation of $E$ must precede $E$ causally), what Pearl called an explanation would be a special case of what we are calling an explanation. However, we allow more general explanations, in that we do not require an explanation to be a truth assignment. In this sense, our framework can be viewed as generalizing Pearl's and Shimony's.

Our definition also borrows heavily from Gärdenfors' definition. We take from him the requirement that $\Pr(X) < 1$. His other requirement, that $\Pr_E^-(E|X) > \Pr_E^-(E)$, will also play a role in our ordering of explanations. The form of the explanation—a conjunction of a (partial) causal mechanism and an instantiations of nodes—is also taken from Gärdenfors.[7] Since we are working with propositional Bayesian networks, the instantiation of nodes clearly corresponds to taking the conjunction of atomic sentences in first-order logic. Gärdenfors allows disjunctions as well (since he allows singular sentences, which are Boolean combinations of atomic sentences). Allowing disjunctions seems to cause problems for us; we return to this issue in Section 4.3. The (partial) causal mechanism can be viewed as a generalization of statistical assertions. We view the requirement of the causal mechanism as a key difference between our definition and Gärdenfors'. For one thing, the causality requirement prevents $E \wedge C$ from being an explanation of $E$, since $E$ cannot precede $E$ in the causal ordering. It also prevents a symptom from being an explanation of a disease.

We would argue that causality is what makes most of Gärdenfors examples involving statistics so compelling. For example, consider the case of Mr. Johansson. We believe that the explanation "70% of those who work with asbestos develop lung cancer" involves more than just the statistical assertion. It is accepted as an explanation because we implicitly accept that there is a causal structure with an edge from a node labeled asbestos to a node labeled lung cancer (with a conditional probability table saying that the probability of lung cancer given asbestos is 0.7). And it is the lack of causality that causes us (if the reader will pardon the pun) not to accept "70% of the time that the barometer reading goes down there is a storm" as an explanation of a storm (unless we happen to believe that barometer readings have a causal influence on storms).

---

[7]Originally, the idea came from Hempel's work on explanation (Hempel and Oppenheim 1948).

However, the situation is different if we try to explain our beliefs to someone else. In this case, the causal structure is symmetric. The fact that I believe that there is a storm does explain my *belief* that the barometer reading has gone down; my belief that the barometer reading has gone down is an explanation for my belief that there is a storm. Ultimately, these beliefs should be rooted in an observation (either of the storm or the barometer).

We can readily convert a causal network describing a situation to a network describing an agent's beliefs. We just reinterpret all the nodes so that a node labeled $X$ talks about the agent's belief in $X$, moralize the graph and change all the directed edges to undirected edges. The resulting *Markov network* (Pearl 1988) captures the causal as well as probabilistic dependencies between the agent's beliefs. Note that the resulting network is no longer asymmetric. While we do not view a symptom as a cause for a disease, believing that a patient has a certain symptom might well cause us to believe that he has a disease. However, an explanation for the agent's beliefs would then be an acyclic subnetwork of this network, together with some new nodes representing the external causes of some of the beliefs. For example, an external cause for the belief that the patient has symptom $d$ is the observation of the symptom; an external cause for the belief that David has an ear infection might be receiving that information from a doctor. We discuss this in more detail in the full paper.

## 4.2 ORDERING EXPLANATIONS

As we have seen in the few examples presented so far, and as is indeed the case in many applications, there are typically several competing explanations. We need to be able to compare them and choose the best. The two proposals presented above for ordering explanations—Gärdenfors' notion of explanatory power and Pearl's notion of considering the probability of the explanation given the explanandum—both have their merits, but neither seems quite right to us. The following example might help clarify the differences between them.

**Example 4.1** *Assume that we have a bag of 100 coins, 99 of which are strongly biased (9:1) towards heads and one that is just as strongly biased towards tails. We pick a coin at random and toss it. The coin lands tails.*

*We can model this situation by using two random variables: $C$ (the type of coin) with values $bh$ and $bt$ (biased towards heads and biased towards tails) and $R$ (the result of the toss), with values $h$ and $t$. A priori, the probability that we picked a coin that is biased towards heads is very high; in fact $P(C=bh) = 0.99$. After receiving the evidence of the coin landing tails, we find out that $P(C=bh|R=t)$ is close to 0.92—less that the prior on $C=bh$ but still very high. What explanation would we accept for the fact that the coin landed tails? Clearly, the causal structure in this situation is known: there is a causal relation between $C$ and $R$, with the obvious conditional probability table described by the story. Since the causal structure is known, the allowable explanations can be identified with $C=bh$ and $C=bt$.*

*What is the relative merit of these explanations?*

According to Gärdenfors' definition, $C = bt$ is a much better explanation than $C = bh$, since $\Pr(R = t|C = bt)$ is much greater than $\Pr(R = t|C = bh)$, where Pr is the prior probability distribution, before the outcome $R = t$ is known. Intuitively, $C = bt$ has far better explanatory power because it accounts for the observation far better than $C = bh$ does. On the other hand, the explanation seems unsatisfactory, since it does not take into account the low probability that the coin biased towards tails will be picked.

According to Pearl's ordering, the best explanation of the coin landing tails is $C = bh$, since $\Pr(C = bh|R = t)$ is much greater than $\Pr(C = bt|R = t)$.[8] This explanation, although very likely itself, doesn't seem to relieve the "cognitive dissonance" between the explanandum and the rest of our beliefs. While it may be the correct *diagnosis* of the situation, it doesn't seem right to call it an *explanation*. The fact that the potential explanation is less probable a posteriori than a priori should at least cause some suspicion.

Notice that by Bayes' rule,

$$\Pr(C = bh|R = t) = \frac{\Pr(R = t|C = bh)}{\Pr(R = t)} \times \Pr(C = bh).$$

The term $\frac{\Pr(R=t|C=bh)}{\Pr(R=t)}$ is what we called the explanatory power of $C = bh$ with respect to $R = t$. Thus, the degree to which $C = bh$ is an explanation of $R = t$ according to Pearl is precisely the product of $EP(C = bh, R = t)$ and the prior probability of $C = bh$. Thus, we can see the precise sense in which Pearl's definition takes into account the prior whereas Gärdenfors' does not.

Although the two definitions disagree in this example, there are many situations of interest in which they agree (which is, perhaps, why both have seemed to be acceptable definitions of the notion of explanation). In particular, they agree in situations where the prior probability of all explanations is the same (or almost the same). Thus, if the user has no particular predisposition to accept one explanation over another, both approaches will view the same explanation as most favorable.[9]

Since we cannot always count on the prior of all explanations being equal, we would like an ordering on explanations that takes into account both the explanatory power and the prior. One obvious way of taking both into account is to multiply them, which is essentially what Pearl does, but multiplication loses significant information and sometimes gets counterintuitive results. (More examples of this appear below.) A straightforward alternative is to associate with each explanation $X$ of $E$ the pair of numbers $(EP(X, E), \Pr_E^-(X))$: the explanatory power of $X$ with respect to $E$ and the prior of $X$. We can then place a partial order $\succeq_E$ on explanations of $E$ by taking $X_1 \succeq_E X_2$ iff $EP(X_1, E) \geq EP(X_2, E)$ and $\Pr_E^-(X_1) \geq \Pr_E^-(X_2)$.

Notice that with this ordering, the two explanations in the coin example, $C = bh$ and $C = bt$, are incomparable. This forces the user to decide whether the explanatory power or the prior is the more significant feature here. In a case like this, such a wide divergence between the explanatory power and the prior of two explanations might signal a problem with the causal model. Perhaps the agent's prior on $C = bh$ vs. $C = bt$ is incorrect in this case.

Although the ordering is partial in general, it can be viewed as a natural generalization of Pearl's ordering. Suppose the causal mechanism is known, as is implicitly assumed by Pearl. If we allow explanations that are complete descriptions of worlds, then all complete descriptions that include $E$ have exactly the same explanatory power: $1/\Pr_E^-(E)$. Thus, our ordering would order them by their prior, just as Pearl's and Shimony's does.

Our ordering also avoids the problem in Gärdenfors' ordering that adding irrelevant conjuncts results in an equally preferred explanation. For example, if $X$ is an explanation of $E$ then $X \wedge Y$ (for all $Y$ conditionally independent of $E$ given the epistemic state) would be considered a worse explanation than $X$ in our ordering since their explanatory powers are the same and $X$'s prior is higher.

If we add a conjunct that is not completely irrelevant, then our approach forces the agent to decide between more specific explanations that have higher explanatory power, and less specific explanations, that have a higher prior. For example, suppose we want to understand why a somewhat sheltered part of the lawn is wet. One possible explanation is that it rained last night, but rain does not always cause that part of the lawn to get wet. A better explanation might be that it was raining and very windy. The combination of rain and wind has better explanatory power than rain alone, but a lower prior. According to our ordering, this makes the two explanations incomparable. This does not seem so unreasonable in this case. We would expect a useful explanatory system to point out both possible explanations, and let the user decide if the gain from the extra explanatory power of wind is sufficiently high to merit the lower prior.

Note that if we multiply the explanatory power of the explanation by its prior, we will always prefer the explanation "rain". To see this, note that for any explanation $X$, the product of the explanatory power and the prior is $\Pr_E^-(X|E)$. Since clearly $\Pr_E^-(X|E) \geq \Pr_E^-(X \wedge Y|E)$, the simpler explanation is preferred. This is a case where multiplication causes a loss of useful information.

### 4.3 DEALING WITH DISJUNCTIONS

As we have defined it, an explanation is a conjunction of a partial causal mechanism together with an instantiation of nodes. We have not allowed disjunctions. Disallowing disjunctions of causal mechanisms seems reasonable. It is consistent with the intuition that "you have cancer either because you smoke or because you have a genetic predisposition to cancer" is viewed as a disjunction of two explanations, not one explanation which has the form of a disjunction. We suspect that it is for similar reasons that Gärdenfors disallowed the disjunction of statistical assertions in his definition.

---

[8] Note that, according to Pearl's definition, $C = bh$ would not be an explanation of $R = t$. The two possible explanations would be $C = bh \wedge R = t$ and $C = bt \wedge R = t$. What we are analyzing here is the ordering produced by Pearl's definition of better explanation on the notion of explanation defined according to our approach.

[9] Here we are also implicitly assuming that there is a prior agreement on what counts as an explanation. As we have observed, the two approaches differ in this respect too.

On the surface, it may seem less reasonable to disallow the disjunction of instantiations of nodes. Certainly it is straightforward to modify our definition so as to allow them, and doing so would be more in keeping with Gärdenfors' allowing singular sentences. However, notice that allowing the disjunction of instantiations has the effect of allowing disjunctions of causal mechanisms.

Consider a case in which we ask for an explanation of huge forest fires recently occurring in California. One possible explanation is that the fire prevention caused the brush to overgrow, another that the tourists often leave campfires unattended. Both these explanations are very plausible (and so is their conjunction). Suppose the agent considers only one causal network possible, and it contains both of these mechanisms. Thus, by allowing the explanation "either some tourists left their campfire unattended or the brush was overgrown", we are effectively allowing a disjunction of causal mechanisms. This example suggests that we may want to make a distinction between what appear to be two different causal mechanisms co-existing within the same causal structure (perhaps using the techniques discussed by Druzdzel and Simon (1993)). This is an area for future research.

On the other hand, there are cases where allowing disjunctions seems useful. For example, consider a situation in which we have four coins, $C_1$, $C_2$, $C_3$, and $C_4$, where $C_1$ and $C_2$, are biased towards heads and $C_3$ and $C_4$ are biased toward tails. We pick one coin at random and toss it three times. The coin lands heads every time. The obvious explanation for this fact is that we picked one of the coins biased towards heads, that is, either $C_1$ or $C_2$. And, indeed, our ordering would prefer the explanation $X_1 =_{def} (C = C_1) \vee (C = C_2)$ to either of the explanations $C = C_1$ or $C = C_2$, assuming that both $C_1$ and $C_2$ had the same bias.

By way of contrast, the explanation $X_2 =_{def} (C = C_1) \vee (C = C_2) \vee (C = C_3)$ does not seem at all reasonable although, according to our ordering, it is incomparable to $X_1$. While $X_1$ has higher explanatory power, $X_2$ has a higher prior. While most people would clearly reject $X_2$, it would be useful to have to have some automatic way of rejecting it.[10]

This example suggests that rather than allowing disjunctions, a better strategy might be to add an additional variable representing the type of the coin (with possible values $bh$ and $bt$, as before). However, we have as yet no principled way for deciding when to add such variables.

Shimony's work can be viewed as an attempt to provide principles as to when to consider disjunctive explanations. The partial explanations of (Shimony 1991) are sets of worlds where the truth values of some primitive propositions are fixed, while the rest can be arbitrary. The sets of partial explanations of (Shimony 1993) correspond to more general sets of worlds, but there are still significant restrictions. For example, the disjunctive explanation must correspond to a node already in the network and the probability of the explanandum must be the same for every disjunct in the disjunctive explanation. The latter restriction is quite severe. In our coin example, if the coins biased towards heads have different biases, Shimony's approach would not allow us to consider the explanation $X_1$, "we picked a coin biased towards heads". Of course, we can easily loosen this restriction to allow disjunctions where the conditional probabilities are *almost* the same. However, it seems to us that we want more than just similar conditional probabilities here. We only want to allow disjunctions if the causal mechanism for each disjunct is the same.

To be fair, Shimony uses his restrictions to allow him to find good explanations algorithmically. It is not clear whether there are also philosophical reasons to restrict them in this way. We hope to explore both the algorithmic and foundational issues in future work.

## 5  CONCLUSIONS

We feel that the contribution of this paper is twofold: First, we present a critique of two important approaches to explanation; second, we outline a sketch of a novel approach that tries to take into account the best features of both, and combine them with a notion of causality.

Our approach clearly needs to be fleshed out. Some areas for future research include:

- Obviously, much of the effort will involve research in causality. Most of the work in causality has allowed only what amounts to propositional reasoning. (The nodes represent random variables that take on a small finite number of values.) Can we extend it to allow causal explanations that involve first-order constructs and temporal constructs? There has been some work on adding these constructs to Bayesian networks (see, for example, (Dean and Kanazawa 1989; Glesner and Koller 1995; Haddawy 1994)), but no work focusing on their application to causal reasoning. Along similar lines, it would be useful to have a good language for reasoning about causality, that allowed first-order reasoning and temporal constructs.

- As we have observed, our approach, which provides only a partial ordering on explanations, seems too weak. While it is not clear that we want to have a total order, it does seem that we want to allow more explanations to be comparable than is the case according to our ordering. This is particularly the case if we allow disjunctive explanations.

---

[10]This is another case where multiplying the components gives a misleading answer: $X_2$ has a higher product than $X_1$. In general, if we compare the explanation $X$ to a disjunctive explanation $X \vee Y$ by multiplying the explanatory power times the prior, then we will always prefer $X \vee Y$ to $X$, for the same reasons as given earlier for preferring $X$ to $X \wedge Y$.

- A natural extension would be to apply our definition to counterfactuals.

  After all, humans seem to have no problem with explaining hypothetical facts. We believe that our basic framework should be able to handle this, although perhaps we may need to use structural equations and the interpretation of counterfactuals given by Balke and Pearl (1994).

- As we said earlier, given that our goal is to have the system provide an explanation that is useful to a user, it would be important to model the user's knowledge state and adjust explanations accordingly. The work of Suermondt (1992) is relevant in this regard. He also puts the emphasis on explaining beliefs (or, specifically, probability distribution over the node of interest) adopted by the system as a result of receiving some observation. His goal is to find a small subset of evidence responsible for this change and the links most influential in transmitting it. In our context, we can understand Suermondt as considering a system which has full knowledge of the domain (characterized by a Bayes Net together with all the conditional probability tables) and knows the values of some variables, trying to explain its beliefs to a user with no (or minimal) knowledge. Thus, for him, an explanation amounts to finding a "small" set of instantiations of variables (i.e., a partial truth assignment) and a 'small" partial causal mechanism that will raise the posterior probability of the observations.

Given the importance of explanation, we believe that these questions represent fruitful lines for further research.

## Acknowledgments

We thank anonymous reviewers for pointing out several important references and Adam Grove for useful discussions. Part of this work was carried out while the second author was at the IBM Almaden Research Center. IBM's support is gratefully acknowledged. The work was also supported in part by the NSF, under grants IRI-95-03109 and IRI-96-25901, and the Air Force Office of Scientific Research (AFSC), under grant F94620-96-1-0323.

## References

Alchourrón, C. E., P. Gärdenfors, and D. Makinson (1985). On the logic of theory change: partial meet functions for contraction and revision. *Journal of Symbolic Logic 50*, 510-530.

Bacchus, F., A. J. Grove, J. Y. Halpern, and D. Koller (1996). From statistical knowledge bases to degrees of belief. *Artificial Intelligence 87*(1-2), 75-143.

Balke, A. and J. Pearl (1994). Counterfactual probabilities: Computational methods, bounds and applications. In *Proc. Tenth Conference on Uncertainty in Artificial Intelligence (UAI '94)*, pp. 46-54.

Boutilier, C. and V. Becher (1995, August). Abduction as belief revision. *Artificial Intelligence 77*, 43-94.

Dean, T. and K. Kanazawa (1989). A model for reasoning about persistence and causation. *Computational Intelligence 5*, 142-150.

Druzdzel, M. J. (1996). Explanation in probabilistic systems: Is it feasible? will it work? In *Proceedings of the Fifth International Workshop on Inteligent Information Systems (WIS-96)*, Dęblin, Poland.

Druzdzel, M. J. and H. A. Simon (1993). Causality in bayesian belief networks. In *Uncertainty in Artificial Intelligence 9*, pp. 3-11.

Gärdenfors, P. (1988). *Knowledge in Flux: Modeling the Dynamics of Epistemic States*. MIT Press.

Glesner, S. and D. Koller (1995). Constructing flexible dynamic belief networks from first-order probabilistic knowledge bases. In *Proceedings of the European Conference on Symbolic and Quantitative Approaches to Reasoning and Uncertainty (ECSQARU '95)*, pp. 217-226.

Haddawy, P. (1994). Generating Bayesian networks from probability logic knowledge bases. In *Proc. Tenth Conference on Uncertainty in Artificial Intelligence (UAI '94)*, pp. 262-269. Morgan Kaufmann.

Halpern, J. Y. (1990). An analysis of first-order logics of probability. *Artificial Intelligence 46*, 311-350.

Heckerman, D. and R. Shachter (1995). Decision-theoretic foundations for causal reasoning. *Journal of Artificial Intelligence Research 3*, 405-430.

Hempel, C. G. (1965). *Aspects of Scientific Explanation*. Free Press.

Hempel, C. G. and P. Oppenheim (1948). Studies in the logic of explanation. *Philosophy of Science 15*.

Henrion, M. and M. J. Druzdzel (1990). Qualitative propagation and scenario-based approaches to explanation of probabilistic reasoning. In *Uncertainty in Artificial Intelligence 6*, pp. 17-32.

Pearl, J. (1988). *Probabilistic Reasoning in Intelligent Systems*. San Francisco, Calif.: Morgan Kaufmann.

Pearl, J. (1995). Causal diagrams for empirical research. Biometrika, 82(4), 669-709, December 1995.

Salmon, W. C. (1984). *Scientific Explanation and the Causal Structure of the World*. Princeton University Press.

Shimony, S. E. (1991). Explanation, irrelevance and statistical independence. In *Proc. National Conference on Artificial Intelligence (AAAI '91)*, pp. 482-487.

Shimony, S. E. (1993). Relevant explanations: Allowing disjunctive assignments. In *Proc. Ninth Conference on Uncertainty in Artificial Intelligence (UAI '93)*, pp. 200-207.

Suermondt, H. J. (1992). *Explanation in Bayesian Belief Networks*. Ph. D. thesis, Stanford University.

Suermondt, H. J. and G. F. Cooper (1992). An evaluation of explanations of probabilistic inference. In *Proc. of the Sixteenth Annual Symposium on Computer Applications in Medical Care*, pp. 579-585.